\documentclass[conference,a4paper]{IEEEtran}
\IEEEoverridecommandlockouts

\usepackage[hidelinks]{hyperref}
\usepackage[cmex10]{amsmath}
\usepackage{amssymb,amsfonts}
\usepackage{dblfloatfix}

\usepackage[ruled,vlined]{algorithm2e}
\usepackage{graphicx}
\graphicspath{{Figures/PDF/}{Figures/PNG/}{Figures/EPS/}}

\usepackage{booktabs}
\usepackage{siunitx}
\usepackage[numbers,compress]{natbib}
\usepackage{texnames}
\usepackage{bm,bbm}
\usepackage{orcidlink}

\usepackage{multirow}

\begin{document}

\title{\uppercase{Stereo Radargrammetry Using Deep Learning from Airborne SAR Images}}

\author{
  \IEEEauthorblockN{Tatsuya Sasayama, Shintaro Ito\orcidlink{0009-0007-8319-5980}, Koichi Ito\orcidlink{0000-0001-7431-7105}, and Takafumi Aoki\orcidlink{0000-0001-8308-2416}}
  \IEEEauthorblockA{\textit{Graduate School of Information Sciences, Tohoku University}\\
  6-6-05, Aramaki Aza Aoba, Sendai, 9808579, Japan.\\
  sasayama@aoki.ecei.tohoku.ac.jp}
}

\maketitle

\begin{abstract}
  In this paper, we propose a stereo radargrammetry method using deep learning from airborne Synthetic Aperture Radar (SAR) images.
  Deep learning-based methods are considered to suffer less from geometric image modulation, while there is no public SAR image dataset used to train such methods.
  We create a SAR image dataset and perform fine-tuning of a deep learning-based image correspondence method.
  The proposed method suppresses the degradation of image quality by pixel interpolation without ground projection of the SAR image and divides the SAR image into patches for processing, which makes it possible to apply deep learning.
  Through a set of experiments, we demonstrate that the proposed method exhibits a wider range and more accurate elevation measurements compared to conventional methods.
\end{abstract}


\section{Introduction}

The information of the ground surface can be acquired by remote sensing with sensors mounted on satellites or aircraft \cite{Richards-Springer-2009}.
Remote sensing has been used for understanding the situation of the affected area, assessing the damage, reducing the effects of natural disasters, and planning rescue operations \cite{Kerle2013}.
To achieve the above, 3D measurement of the ground surface is an indispensable technique in remote sensing.

A Synthetic Aperture Radar (SAR) has been used in remote sensing, which can acquire images of the ground surface by emitting radio waves and receiving their reflected waves with antennas mounted on a moving platform such as a satellite or an airplane.
The advantages of using SAR in remote sensing are independence from weather and time of day, pulse compression of SAR, and aperture synthesis to achieve extremely high spatial resolution.
Therefore, we have been investigating methods of 3D measurement using SAR amplitude images \cite{Maruki,hishinuma,hishinuma2,Karl}.
Note that the SAR amplitude image is referred to as a SAR image in the following.

We focus on radargrammetry, which can measure the elevation of the target area by image correspondence between two SAR images taken on different flight paths \cite{maitre}.
In our previous work \cite{Maruki,hishinuma,hishinuma2,Karl}, to take advantages of the flexibility of flight path and the absolute elevation measurement of radargrammetry and to achieve the comparable measurement accuracy with Interferometric SAR (InSAR) \cite{Rosen-IEEE-2000}, we have introduced techniques of stereo vision to radargrammetry, i.e., stereo radargrammetry.

Stereo radargrammetry is to measure the absolute elevation of the ground surface based on the principle of stereo vision from the correspondence between the SAR images acquired at two different paths.
In our previous work \cite{Maruki,hishinuma,hishinuma2,Karl}, we have employed Phase-Only Correlation (POC) \cite{POC} for obtaining correspondence between the two SAR images.
POC is the image matching technique using phase information obtained by discrete Fourier transform of given images \cite{POC}.
The correspondence matching method using POC employs a coarse-to-fine strategy using image pyramids for robust correspondence search and a sub-pixel translational displacement estimation method using POC for local block matching.
On the other hand, POC cannot always achieve accurate correspondence between SAR images due to geometric image modulation inherent in SAR, such as distortions and invisible regions.
In the field of computer vision, image correspondence methods using deep learning \cite{DKM,RoMa} have been proposed, which achieve higher accuracy than image matching-based methods.
In particular, these methods are considered to suffer less from geometric image modulation since they can provide accurate correspondence even between images with widely different viewpoints.
However, deep learning-based methods using are not necessarily suitable for SAR images since they are trained on a large number of camera images.

In this paper, we create a SAR image dataset, which is necessary to use image correspondence methods using deep learning for stereo radargrammetry, and propose a novel stereo radargrammetry method with deep learning-based image correspondence fine-tuned on our dataset.
The proposed method employs Robust Dense Feature Matching (RoMa) \cite{RoMa}, which is a state-of-the-art deep learning-based image correspondence method.
Our dataset used for training and evaluation of deep learning-based methods consists of SAR images acquired around Mt. Aso in Kumamoto, Japan and corresponding Digital Surface Models (DSM) \cite{DSM}.
Through a set of experiments, we demonstrate that the proposed method achieves higher accuracy of stereo radargrammetry than the POC-based method and another deep learning-based image correspondence method.

\begin{table}[t]
  \caption{Specification of SAR image dataset used in this paper.}
  \label{tbl:dataset_split}
  \centering
  \begin{tabular}{cccccc}
    \hline
    \multirow{2}{*}{Method} & \multirow{2}{*}{Size of patch images} & \multicolumn{4}{c}{\# of patch images} \\
    \cline{3-6}
     & & Train & Val & Test & Total\\
    \hline
    DKM \cite{DKM} & 384$\times$512 pixels & 3,001 & 1,019 & 372 & 4,392 \\
    RoMa \cite{RoMa} & 560$\times$560 pixels & 1,656 & 552 & 207 & 2,415 \\
    \hline
  \end{tabular}
\end{table}
\begin{figure}[t]
    \centering
    \includegraphics[width=\linewidth]{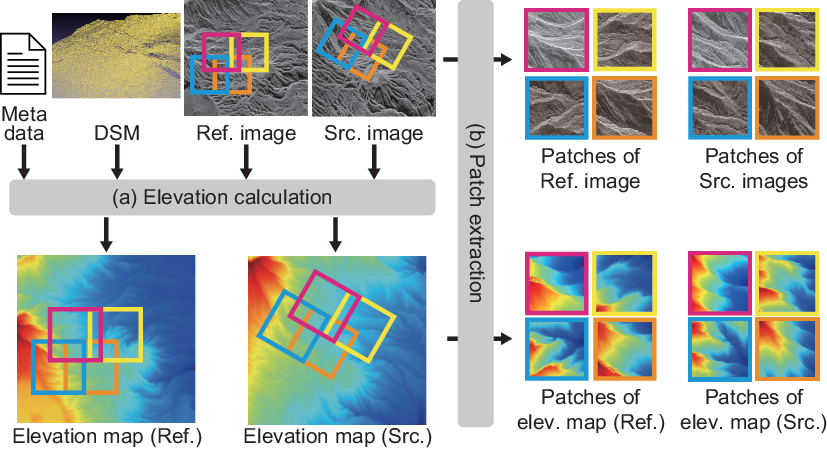}
    \caption{Overview of creating the SAR image dataset for applying image correspondence methods using deep learning to stereo radargrammetry.}
    \label{fig:dataset-creation}
\end{figure}

\begin{figure*}[t]
  \centering
  \includegraphics[width=0.9\linewidth]{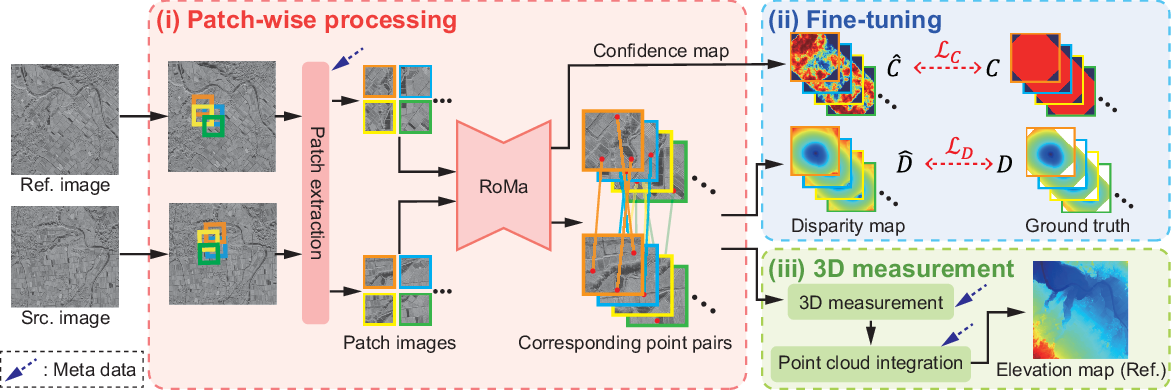}
  \caption{Overview of the proposed method consisting of (i) patch-wise processing, (ii) fine-tuning, and (iii) 3D measurement.}
  \label{fig:stereo_radagrametory}
\end{figure*}

\section{Stereo Radargrammetry Using Deep Learning}

This section describes the details of stereo radargrammetry using deep learning, which is proposed in this paper.

\subsection{Overview of Stereo Radargrammetry}
\label{sec:stereo}

We leverage the stereo radargrammetry framework of our previous work \cite{Karl}, which can measure the elevation from two SAR images by stereo vision.
The internal and external parameters are obtained from the metadata in SAR image acquisition and the SAR projection model takes the earth ellipsoid into account.
For more details, please refer to \cite{Karl}.
The major differences from our previous work \cite{Karl} are as follows: (i) RoMa \cite{RoMa} is used for image correspondence, (ii) the SAR image dataset is created for utilizing deep learning in SAR image processing, and (iii) ground projection is not applied to SAR images.
In (iii), deformation between images has to be approximated as translations by ground projection to use POC \cite{POC}, resulting in degradation of image quality due to pixel interpolation in the conversion from slant-range images to ground-range images.
The proposed method directly inputs the slant-range images, since RoMa \cite{RoMa} allows image correspondence between images with complex deformation.

\subsection{Dataset Creation for Deep Learning}
\label{sec:SAR_Dataset}

Image correspondence methods using deep learning assume that camera images are input, and do not assume that SAR images are input.
In addition, the SAR image dataset which required to train the image correspondence methods using deep learning has not been publicly available.
Therefore, we create the dataset to apply deep learning-based methods to stereo radargrammetry.
The dataset consists of SAR image pairs and their corresponding elevation map pairs.
SAR images acquired on April 17, 2016 and November 16, 2017 using the airborne X-band SAR system (Pi-SAR2) \cite{NICT_sar}, which is developed by National Institute of Information and Communications Technology (NICT), Japan, around Mount Aso in Kumamoto Prefecture, Japan.
We selected 25 and 15 pairs from the SAR images acquired in 2016 and 2017, respectively, and use 24 pairs for training, 8 pairs for validation, and 3 pairs for test, where SAR image pairs consists of two SAR images with the same observation area.
Note that the remaining 5 pairs cannot be included in the dataset since they have the same observation area as the test dataset.
To prepare the elevation map of each observation area, we use DSM \cite{DSM} with a resolution of 0.5 m/pixel, which is measured in the same area using the WorldView satellite of Maxar, Inc.
SAR images are divided into patches of trainable size as shown in Fig. \ref{fig:dataset-creation}, since the size of the SAR image is approximately $8,000 \times 8,000$ pixels.
First, the elevation map corresponding to the SAR image pair is obtained from DSM using the metadata and the projection model as shown in Fig. \ref{fig:dataset-creation} (a).
Next, patches are extracted from the SAR image pair as shown in Fig. \ref{fig:dataset-creation} (b).
A patch is defined on the reference (Ref.) image and extracted from the corresponding position on the source (Src.) image using the latitude and longitude in the metadata.
Note that the patches on Ref. image should be selected so that 1/3 of the image overlaps between adjacent patches so that disparity can be calculated from a patch pair.
A patch is also extracted from the elevation map based on the location of the patch on Ref. image.
The size of the patch is defined according to the input size of image correspondence methods, e.g., $384 \times 512$ pixels for DKM \cite{DKM} and $560 \times 560$ pixels for RoMa \cite{RoMa}.
For each SAR image pair, about 120 and 70 patch pairs are extracted from patches of $384 \times 512$ pixels and $560 \times 560$ pixels, respectively.
The specification of the created dataset is shown in Table \ref{tbl:dataset_split}.

\subsection{Proposed Method}
\label{sec:proposed}

We describe the detail of the proposed method, which introduces RoMa \cite{RoMa} to stereo radargrammetry.
RoMa extracts multi-scale features from input image pairs using DINOv2 \cite{Dino-v2} and VGG-19 \cite{VGG-19}, and estimates the global deformation between images using rough image features.
The fine image features are used to refine the deformation between images, and then the dense correspondence between images are obtained using the refined deformation between images.
RoMa is expected to be effective for finding correspondence between SAR images because of its high accuracy for large deformation and poor-texture regions between camera images.
As mentioned above, not only replacing the image correspondence part of our conventional method \cite{Karl} with RoMa \cite{RoMa}, the proposed method divides the SAR image into patches for image correspondence, and fine-tunes RoMa using the SAR images.
Fig. \ref{fig:stereo_radagrametory} illustrates the overview of the proposed method, which consists of (i) patch-wise processing, (ii) fine-tuning, and (iii) 3D measurement.

\noindent
{\bf (i) Patch-wise processing} --- 
As in Sect. \ref{sec:SAR_Dataset}, the input images are divided into patches based on the latitude and longitude in the metadata.
The patch image pairs that correspond in latitude and longitude between the Ref. and Src. images are input to RoMa and the corresponding point pairs are obtained for each patch pair.
When fine-tuning RoMa with SAR images, a confidence map that visualizes the confidence in the correspondence is also obtained.

\noindent
{\bf (ii) Fine-tuning} --- 
For each patch pair in the training data, RoMa is used to obtain the corresponding point pairs and their confidence maps $\hat{C}$, and the disparity map $\hat{D}$ between the patch images is obtained from the difference in the coordinates of the corresponding point pairs.
Let $D$ and $C$ be the ground truth of the disparity map and the confidence map, respectively.
$D$ is calculated from the difference of the elevation maps of the ground truth, and $C$ is obtained by setting the pixels for which disparity is calculated in $D$ to 1 and the other pixels to 0.
In fine-tuning, the regression loss $\mathcal{L}_D$ and the binary cross-entropy loss $\mathcal{L}_C$ are used.
$\mathcal{L}_D$ is a loss function based on pixel errors between the disparity maps, which is defined by
\begin{equation}
  \label{eq:LW}
  \mathcal{L}_D = \frac{1}{|\bm{I}|} \sum_{i \in \bm{I}} C_i \| \hat{D}_i - D_i \|_2,
\end{equation}
where $\bm{I}$ is a set of pixels in Ref. image, and $\hat{D}_i$, $D_i$, and $C_i$ are the value of $\hat{D}$, $D$, and $C$ for each pixel $i$, respectively.
$\mathcal{L}_C$ is a loss function based on errors between the confidence maps, which is defined by
\begin{equation}
  \label{eq:LC}
  \mathcal{L}_C = \frac{1}{|\bm{I}|} \sum_{i \in \bm{I}} \left\{ C_i \log \hat{C}_i + (1 - C_i) \log (1 - \hat{C}_i) \right\},
\end{equation}
where $\hat{C}_i$ is a value of $\hat{C}$ for each pixel $i$.
The total loss function $\mathcal{L}$ used in fine-tuning is given by
\begin{equation}
  \label{eq:L}
  \mathcal{L} = \mathcal{L}_D + \lambda\mathcal{L}_C,
\end{equation}
where $\lambda$ is a weight.

\noindent
{\bf (iii) 3D measurement} --- 
As in \cite{Karl}, the 3D coordinates are calculated from the corresponding point pairs obtained from each patch pair using the projection model and metadata.
The 3D point cloud obtained for each patch pair is integrated to obtain a 3D point cloud of the target area.
Finally, an elevation map of the target area is obtained from the 3D point cloud.
Note that the elevation is not calculated for the pixels on Ref. image that cannot be matched, such as outliers.

\section{Experiments and Discussion}

We demonstrate the effectiveness of the proposed method through the following experiments.

\begin{figure}[t]
  \centering
  \includegraphics[width=0.85\linewidth]{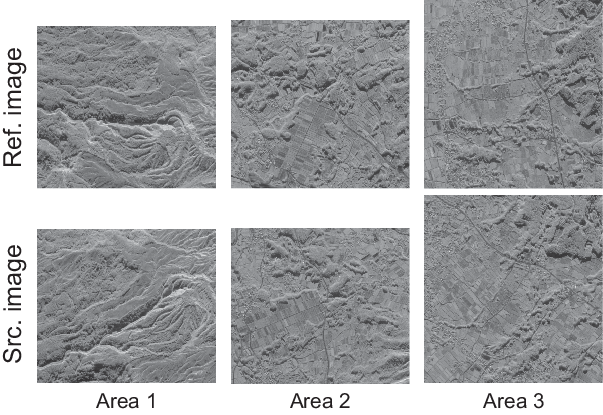}
  \caption{SAR images in the test dataset of Table \ref{tbl:dataset_split}, whose observation area is 2 km$\times$2 km and the intersection angle is about 43 degrees.}
  \label{fig:sar_image}
\end{figure}
\begin{figure*}[t]
    \centering
    \includegraphics[width=0.95\linewidth]{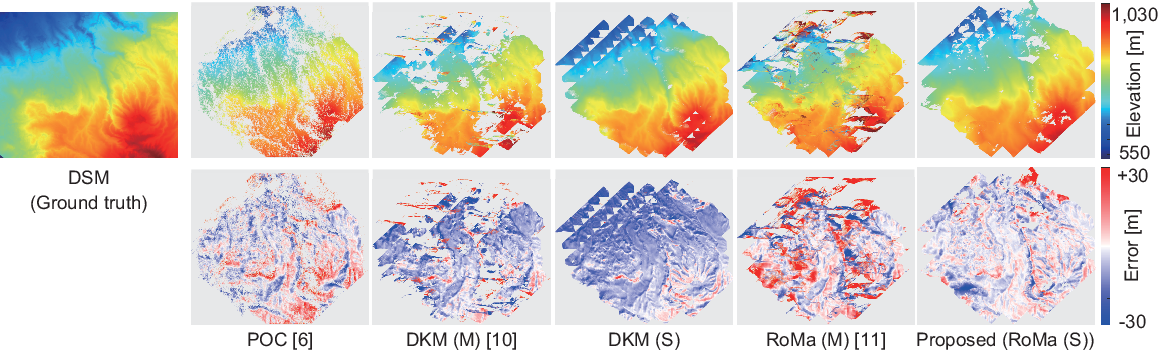}
    \caption{Elevation maps and error maps obtained from each method for Area 1.}
    \label{fig:qualitative-result}
\end{figure*}
\begin{figure}[t]
    \centering
    \includegraphics[width=\linewidth]{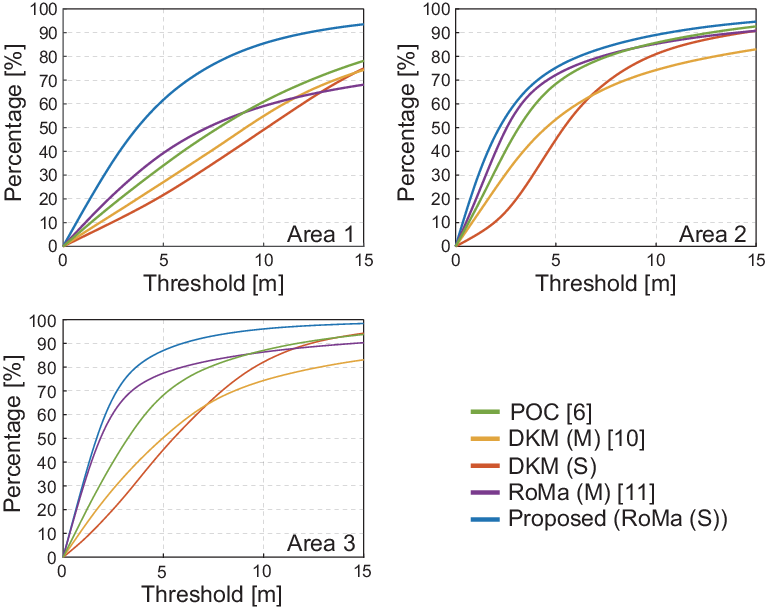}
    \caption{Percentages of 3D points less than the thresholds for errors.}
    \label{fig:error_graph}
\end{figure}
\begin{table*}[t]
  \centering
  \caption{Experimental results for each method, where the values indicate the average error [m] and the standard deviation of errors [m], and the values in parentheses indicate the percentage [\%] of 3D points with an error of 2 m or less.}
  \label{tbl:result_teiryo_mean}
  \begin{tabular}{rcccccccccc}
    \hline
    Dataset & \multicolumn{2}{c}{POC \cite{Karl}}  & \multicolumn{2}{c}{DKM (M) \cite{DKM}} & \multicolumn{2}{c}{DKM (S)} & \multicolumn{2}{c}{RoMa (M) \cite{RoMa}} & \multicolumn{2}{c}{Proposed (RoMa (S))} \\
    \hline
    Area 1  
    & \textbf{0.90} $\pm$ 21.24 & (21.14)  
    & $-5.33$ $\pm$ 39.97 & (16.20)  
    & $-9.83$ $\pm$ \textbf{8.41} & (12.46) 
    & 14.40 $\pm$ 73.29 & (16.20) 
    & $-1.24$ $\pm$ 9.94 & (\textbf{42.20}) \\ 
    Area 2  
    & \textbf{-0.56} $\pm$ 9.89 & (47.88)  
    & $-5.59$ $\pm$ 19.90 & (36.11)  
    & $-5.96$ $\pm$ 6.96 & (19.58)  
    & 2.15 $\pm$ 20.81 & (56.50)  
    & $-1.62$ $\pm$ \textbf{6.52} & (\textbf{60.52}) \\
    Area 3  
    & \textbf{-0.14} $\pm$ 12.12 & (47.11) 
    & $-4.50$ $\pm$ 32.55 & (33.60) 
    & $-5.60$ $\pm$ 5.61 & (25.03) 
    & 2.37 $\pm$ 28.91 & (66.47) 
    & $-1.56$ $\pm$ \textbf{4.28} & (\textbf{74.10}) \\
    \hline
  \end{tabular}
\end{table*}

\subsection{Experiments}

In the experiments, we evaluate the accuracy of the proposed method using three SAR image pairs of the test dataset as shown in Fig. \ref{fig:sar_image}.
Note that the observation areas of the three SAR images are not included in the train and validation datasets.
Area 1 is a mountainous area with large elevation changes, Area 2 has many areas hidden by trees, and Area 3 is a plain area with small elevation changes.
In this experiment, we demonstrate the effectiveness of RoMa \cite{RoMa} in stereo radargrammetry compared to DKM \cite{DKM}.
We also compare the proposed method with the case trained with MegaDepth \cite{MegaDepth}, which is a camera image dataset generally used in deep learning-based methods, to demonstrate the effectiveness of fine-tuning using SAR images.
The models trained on MegaDepth are denoted as DKM (M) and RoMa (M), and the models fine-tuned on the SAR image dataset are denoted as DKM (S) and RoMa (S).
For fine-tuning, the learning rate of the encoder is $10^{-6}$, the learning rate of the others is $2 \times 10^{-5}$, the batch size is 2, $\lambda$ is 0.01, and the number of epochs is 31.
We compare the accuracy with that of the POC-based method \cite{Karl} to demonstrate the effectiveness of the image correspondence method using deep learning.
The measurement accuracy is evaluated by the error between the elevation map obtained by each method and the ground truth, and the percentage of 3D points whose error is less than a threshold.
Note that the external parameters are optimized by aligning DSM with the results of the POC-based method \cite{Karl} to minimize the influence of errors in the external parameters.

\subsection{Results and Discussion}

Table \ref{tbl:result_teiryo_mean} shows a summary of the measurement accuracy of each method, Fig. \ref{fig:qualitative-result} shows the elevation map and error map of each method for Area 1, and Fig. \ref{fig:error_graph} shows the percentage of 3D points less than the thresholds for errors.
In Fig. \ref{fig:qualitative-result}, the gray area indicates pixels for which no 3D points are measured.
The POC-based method \cite{Karl} has the smallest average error but the large standard deviation.
The small average error is due to the cancellation of positive and negative errors, and therefore the standard deviation of errors is important in the quantitative evaluation.
The percentage of 3D points measured less than an error of 2 m is small, indicating that many corresponding points cannot be obtained due to image modulation.
DKM (M) \cite{DKM} and RoMa (M) \cite{RoMa} have large average errors and standard deviations, indicating that they are not adapted to SAR images.
DKM (S) can measure a wider area than DKM (M) \cite{DKM}, while the average error is not improved.
On the other hand, the proposed method, RoMa (S), has the small average error and standard deviation, and the largest percentage of 3D points less than an error of 2 m.
Therefore, RoMa fine-tuned with the SAR image dataset is effective for stereo radargrammetry to measure a wide area with high accuracy.

\section{Conclusion}

In this paper, we created the SAR image dataset for image correspondence methods using deep learning, and proposed a novel stereo radargrammetry method with RoMa, which is a deep learning-based image correspondence method and is fine-tuned on our dataset.
Through a set of experiments, we demonstrated that the proposed method achieves higher accuracy of stereo radargrammetry than the POC-based method \cite{Karl} and the DKM \cite{DKM}-based method.

\section{Acknowledgment}

This work was supported in part by JSPS KAKENHI JP 23H00463 and 25K03131.

{\small
  \bibliographystyle{IEEEtranN}
  \bibliography{references}
}

\end{document}